\documentclass{article}

\usepackage{arxiv}

\usepackage[utf8]{inputenc} % allow utf-8 input
\usepackage[T1]{fontenc}    % use 8-bit T1 fonts
\usepackage{hyperref}       % hyperlinks
\usepackage{url}            % simple URL typesetting
\usepackage{booktabs}       % professional-quality tables
\usepackage{amsfonts}       % blackboard math symbols
\usepackage{nicefrac}       % compact symbols for 1/2, etc.
\usepackage{microtype}      % microtypography
\usepackage{lipsum}		% Can be removed after putting your text content
\usepackage{graphicx}
\usepackage{natbib}
\usepackage{doi}

\usepackage{titlesec} 
\usepackage{mathtools}
\usepackage{breqn}

\title{Toolpath design for additive manufacturing using deep reinforcement learning}

\author{ \hspace{1mm}Mojtaba Mozaffar \\
	Department of Mechanical Engineering\\
	Northwestern University\\
	Evanston, IL 60208, USA \\
	\texttt{mozaffar@u.northwestern.edu}
	\And
	\hspace{1mm}Ablodghani Ebrahimi \\
	Department of Industrial Engineering and Management Sciences\\
	Northwestern University\\
	Evanston, IL 60208, USA \\
	\texttt{ghani@u.northwestern.edu}
	\AND
	\hspace{1mm}Jian Cao\thanks{Corresponding author} \\
	Department of Mechanical Engineering\\
	Northwestern University\\
	Evanston, IL 60208, USA \\
	\texttt{jcao@northwestern.edu}
}

\renewcommand{\headeright}{}
\renewcommand{\shorttitle}{Toolpath design for additive manufacturing using deep reinforcement learning}

\hypersetup{
pdftitle={Toolpath design for additive manufacturing using deep reinforcement learning},
pdfsubject={},
pdfauthor={Mojtaba Mozaffar, Ablodghani Ebrahimi, Jian Cao},
pdfkeywords={Additive Manufacturing, Toolpath, Deep Learning, Reinforcement Learning},
}

\titleformat{\subsubsection}[runin]
  {\normalfont\normalsize\bfseries}{\thesubsubsection}{1em}{}

\begin{document}
\maketitle

\begin{abstract}
Toolpath optimization of metal-based additive manufacturing processes is currently hampered by the high-dimensionality of its design space. In this work, a reinforcement learning platform is proposed that dynamically learns toolpath strategies to build an arbitrary part. To this end, three prominent model-free reinforcement learning formulations are investigated to design additive manufacturing toolpaths and demonstrated for two cases of dense and sparse reward structures. The results indicate that this learning-based toolpath design approach achieves high scores, especially when a dense reward structure is present.
\end{abstract}

% keywords can be removed
\keywords{Additive Manufacturing \and Toolpath \and Reinforcement Learning \and Deep Learning}

\section{Introduction}
Additive Manufacturing (AM) processes offer unique capabilities to build low-volume parts with complex geometries and fast prototyping from a variety of materials. Metal-based AM has become increasingly more popular over the last decade for manufacturing and repairing functional parts in automotive, medical and aerospace industries. Despite the great potential in metal-based AM market, the state-of-the-art practices involve rigorous trial and errors before achieving consistent parts with the desired geometric and material properties, which is mainly due to the sensitivity of the build on process parameters. While the influence of process parameters such as laser power, powder parameters, and scan speed on the microstructure and final properties of the AM build are extensively studied in the literature, the influence of toolpath strategies yet to be fully investigated. 

Authors in \citep{steuben2016implicit} considered three different toolpath patterns for building a part using a fused deposition modeling process and demonstrated that the pattern has a significant effect on the ultimate strength and elastic modulus of the build. Akram et al. \citep{{akram2018understanding}} formulated a microstructure model using a Cellular Automata (CA) and demonstrated a strong correlation between the toolpath pattern (i.e., unidirectional and bi-directional) and the grain orientations. In \citep{bhardwaj2018effect}, the authors considered bi-directional and cross-directional toolpath strategies to manufacture cubic parts with a Direct Metal Laser Sintering (DMLS) process and studied the surface finish, residual stress and mechanical properties of the parts. Their study indicates that the parts built with cross-directional strategy display better mechanical properties, which is due to their desirable microstructure of columnar cells.

From the above-mentioned research, it can be evidently seen that the choice of the toolpath greatly influences various properties of AM builds. However, existing research does not offer a robust solution for the analysis of this influence nor tools to prudently design toolpaths. In this work, we present a novel way to represent the toolpath and learn design strategies that lead to optimal performance. The toolpath design is proposed to be modeled as a Reinforcement Learning (RL) problem in which an agent learns to design optimal toolpaths as it dynamically interacts and collects data from an additive manufacturing environment.

RL is a subfield of artificial intelligence, which focuses on training agents that can interact with an environment and maximize the rewards that the agent collects through this interaction. In an RL schema, the agent is responsible for determining the actions at each time step ($a_t$), which influences the environment causing it to move from its current state ($s_t$) to a new state ($s_{t+1}$) and generates a reward feedback ($r_t$) for the agent. Here, “state” refers to the representation of the environment that is visible to the agent. The agent learns to maximize the long-term rewards that it receives in its lifespan by attempting more of the strategies that lead to the most favorable rewards.

Most modern RL algorithms can be categorized into three main classes: 
\begin{enumerate}
	\item \textbf{Policy optimization methods } parameterize the policy, $\pi_\theta (a|s)$, and optimize $\theta$ to maximize the expected reward. This class of RL algorithms often suffer from poor sample efficiency, requiring millions of samples. Furthermore, for most policy optimization algorithms, all samples should be generated using the agent’s policy at each training step, which exacerbates the sample efficiency of these methods as historic data cannot be used in the training process. 
	\item \textbf{Value function optimization methods}  (also called Q-learning) do not optimize the policy directly. Rather, they aim to find the optimal action-value function $Q^* (s,a)$, as defined in Eq. \ref{eq:1}, to represent the maximum discounted reward the agent can collect from any state. Following the actions that lead to maximum optimal action-values, $a^* = argmax_a Q^* (s,a)$, guides the agent to maximize its reward.
	\begin{equation} \label{eq:1}
	Q^* (s,a)= max_\pi \mathbb{E}\left[\sum _{t=0}^H \gamma^t r_t |\pi,s_0=s ,a_0=a \right]
	\end{equation}
	In Eq. \ref{eq:1}, $\pi$ represents the policy, $H$ represents the environment horizon and $\gamma$ is the discount factor---a positive value smaller than 1 . The action-value function can be obtained using the Bellman equation (Eq. \ref{eq:2}) with guaranteed convergence in tabular cases. By exploiting the self-consistency of the problem structure through the Bellman equation, action-value function optimization methods can learn the optimal action-value function and implicitly determine the policy with fewer samples. However, Q-learning lacks the stability of policy optimization methods.
	\begin{equation} \label{eq:2}
	Q^* (s_t,a_t )=r(s_t,a_t )+ \gamma max_{a_{t+1}} Q^* (s_{t+1},a_{t+1} )
	\end{equation}
	\item \textbf{Model-based RL algorithms}, unlike the first two categories (known as model-free RL), attempt to learn an explicit model of the underlying dynamics of the environment. The model is further used for look-ahead planning or as a virtual sample generator. This class of solutions can offer a great sample efficiency with orders of magnitude less required data. However, the quality of the RL agent heavily depends on the accuracy of the dynamics model. Therefore, while there are many successful examples of this approach for robotics and games with perfect environments such as chess, the state-of-the-art algorithms in this class fail in high-dimensional spaces (e.g., pixel-level visual inputs) and uncertain environments.
\end{enumerate}

Note that these categories are not mutually exclusive. In fact, most of the successful existing literature uses a combination of these approaches. Most famously, actor-critic methods simultaneously parametrize and train both policy and value functions.  For example, A2C \citep{mnih2016asynchronous} follows the policy gradient theorem while using the value function to reduce the variance of gradient estimation and provides a stable solution for continuous action spaces.

In this work, we investigate a number of leading model-free candidates that showed promising results in domains such as Atari games. Investigation of model-based approaches is not considered here because of their often-suboptimal performance in high-dimensional domains such as ours. Our toolpath design system allows exploring an unknown dynamic physics through experiments and opens new avenues for high dimensional design in manufacturing processes.

\section{Methods}
\subsection{AM virtual environment}
\begin{figure}
	\centering
	\includegraphics[width=14cm]{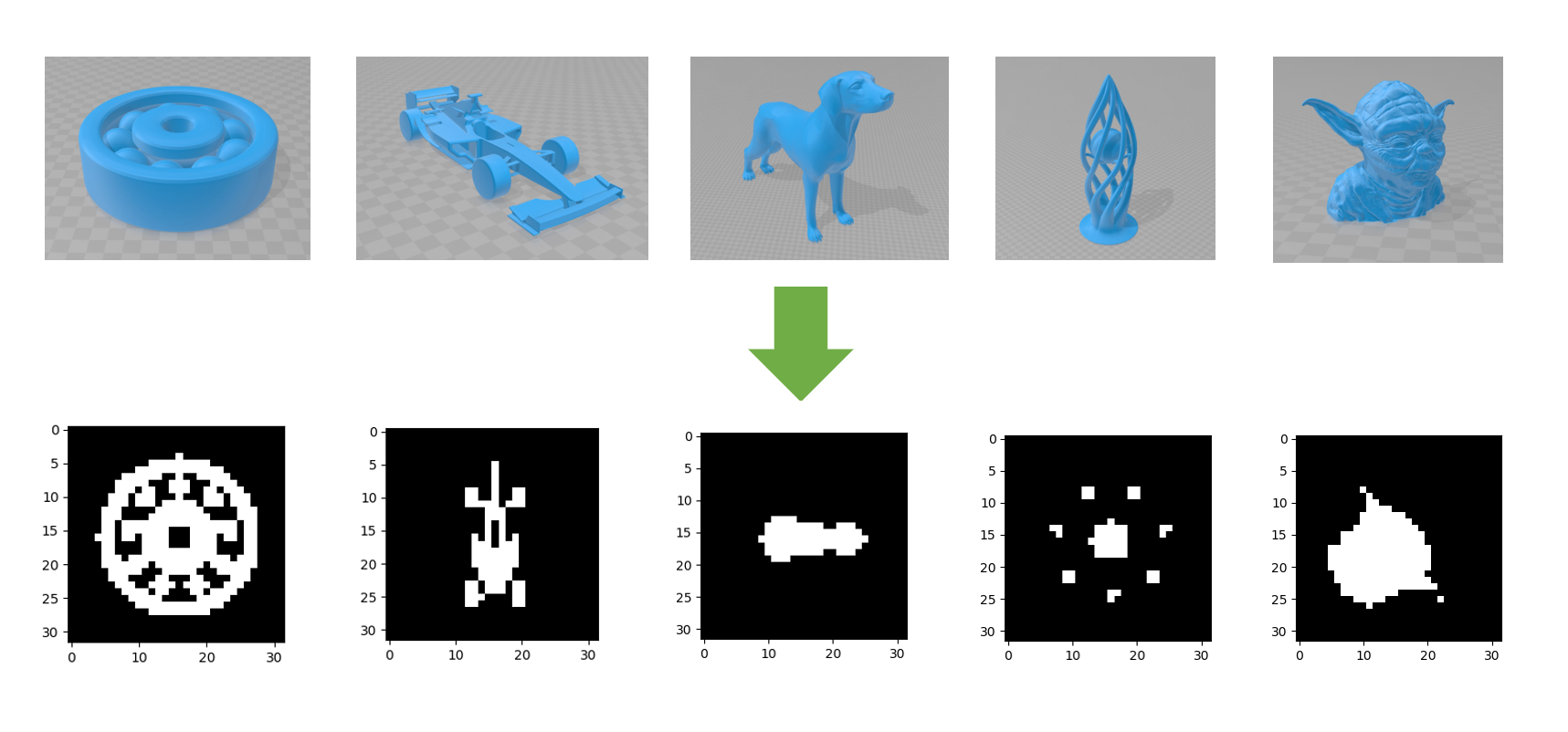}
	\caption{Sample CAD geometries (top row) and pixelized two-dimensional sections (bottom row) for the AM virtual environment.}
	\label{fig:fig1}
\end{figure}

We develop a virtual environment of an AM process, resembling Directed Energy Deposition (DED) processes, to collect data and perform the training process. The virtual environment considers two-dimensional sections on which materials need to be deposited. As we want the strategies learned by the RL agent to be geometry-agnostic, we develop a database of CAD geometries representing a wide range of spatial structures. The CAD geometries are then processed into multiple sections by slicing them in different heights and converted into over 400 two-dimensional sections, each with 32x32 pixels, to train the agent. A sample of considered CAD geometries acquired from Thingiverse online repository \citep{thingiverse.com} and one of their corresponding sections are demonstrated in Fig. \ref{fig:fig1}.

While evaluating the virtual environment, one section is randomly selected and the RL agent is asked to design the toolpath for it one action at a time. Eight actions are available for the agent to explore, including four directions of motion each with two deposition status (on/off). The environment keeps track of the desired section, filled section, and the location and status of the nozzle. A representation of the environment state space ($s_t$) is accessible to the agent at each time step. Once the agent finishes its assigned task for a section (e.g., depositing material on all pixels of the desired section), a new section is randomly selected, and the agent is asked to start over. To avoid excessively long episodes of training on one section, a maximum of 400 actions are selected for each section.

\begin{figure}
	\centering
	\includegraphics[width=10cm]{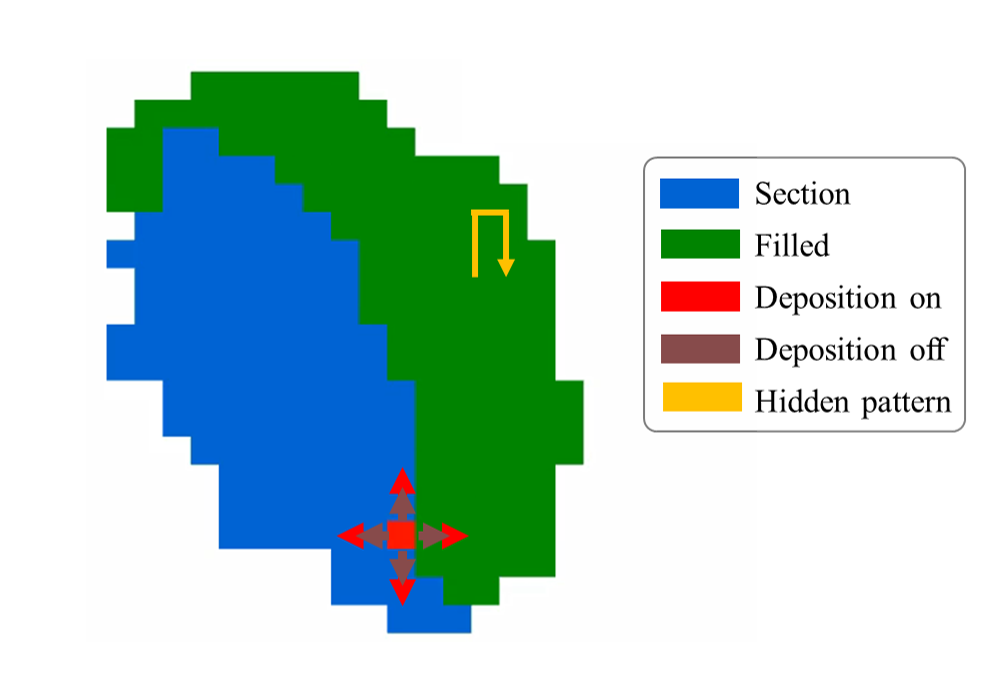}
	\caption{AM virtual environment including section (in blue), filled partition (in green), and nozzle location and status. The red point indicates the location of the nozzle with “on” status. Valid actions are shown with eight arrows for “on” (red) and “off” (brown) status and four directions. The hidden pattern for sparse reward system is plotted in yellow.}
	\label{fig:fig2}
\end{figure}

As can be seen from Fig. \ref{fig:fig2}, we assume a pixelized section representation and discrete action spaces. These two assumptions are not inherently restrictive for the proposed methodology as the representation of the section can be replaced with other continuous or discrete heuristics and the action space can be easily extended to higher number of option (e.g., 8 or 16 directions) or continuous action spaces with minimal change to the algorithm.

We design the state representation ($s_t$) as a single-channel two-dimensional image, where the unfilled section has a value of 1.0 and the rest of the pixels are zero. Additionally, we provide the network with a one-hot encoded list of 10 most recent action histories. The image is first processed through three layers of convolutional neural network, then concatenated with the action history input, followed by two fully connected neural network layers for policy and value networks in each algorithm.

\subsection{Analysis cases}\label{sec:alanysis_cases}
We consider two scenarios for the tasks and their corresponding reward systems in this study:
\begin{enumerate}
\item \textbf{Dense reward system:} In this analysis, we consider a scenario in which a reward can be assigned based on the interaction of agents and environments at each time step. Designing a toolpath that deposits material in all desirable locations of the section in optimal time is an example of a dense reward system. In this case, we assign a reward of 1.0 to any desirable material deposition, -1.0 to material deposition in incorrect locations, and -0.5 to motions without deposition to provide an incentive to finish the toolpath in the shortest time. It is noteworthy that dense reward structures are not limited to static properties of the environment, such as finishing the toolpath. Other examples of dense reward structures include rewards that are assigned based on the meltpool size or shape from an online thermal imaging system.
content...
\item \textbf{Sparse reward system:} As many interesting AM process parameters can only be measured and evaluated after the part is made, a reward can only be assigned to the completed toolpath at the last time step of the episode, which results in a sparse reward system. To simulate this scenario in the developed virtual environment, we consider the sequence of ordered actions (up, up, right, down, and down in this order, see Fig. \ref{fig:fig2}) as a potential desirable pattern and assign a reward at the end of each episode of the simulation based on the similarity of the toolpath generated by the agent with the selected pattern. Note that this pattern is completely hidden from the agent, i.e., the agent can only interpret the pattern through the sparse reward it receives at the end of each episode. The similarity between the toolpath and the hidden pattern is measured by counting the occurrence of the completed or partially completed (with a minimum of three consecutive actions) hidden patterns in the toolpath history. To encourage the agent to finish the toolpath while learning the hidden pattern, an auxiliary reward of 0.1 and -0.1 is assigned for correct and incorrect material deposition respectively. While this specific sequence is selected as a demonstration in this work, the formulation does not depend on it, and the reward structure can be based on any unknown physics of the environment.
\end{enumerate}

\subsection{RL algorithms}
In this work, we mainly focus on investigating and improving three state-of-the-art RL algorithms that have shown to be successful across many tasks. 

\subsubsection{Deep Q-network (DQN)} 
\citep{mnih2015human}, is a Q-learning approach that parametrizes action-value function, Q(s,a), using neural network and iteratively solves the Bellman equation (Eq. \ref{eq:2}) while using a number of numerical techniques to overcome problems associated with training neural networks in RL non-stationary setting. As neural networks generalize, the Bellman equation (Eq. \ref{eq:2}) tails a dynamic target (i.e., both $Q^* (s_t,a_t )$ and $Q^* (s_{t+1},a_{t+1} )$ change while training), which impedes the training process. DQN uses an additional neural network as a target network to estimate the action-value for future states $Q^* (s_{t+1},a_{t+1} )$ and solve the Bellman equation in a more supervised fashion. While the neural network training theories stand on the assumption of independent and identically distributed (i.i.d.) data, the successive data collected in RL settings are greatly correlated. To overcome this issue, DQN uses a replay buffer that stores all transactions of the environment and randomly draw samples from them during the training process. DQN uses epsilon-greedy strategy for exploration, in which the agent initially explores while gradually over the training process learns to acts more according to its predicted model.

In this work, we consider a variation of the original DQN paper that empirically showed enhanced performance for this application. A corrected replay buffer, as proposed in \citep{zhang2017deeper}, is used where the last added sample into replay buffer will be added to the randomly selected batch to eliminate the need for excessively large replay buffer. To reduce the overestimation bias of Q value caused by the maximizing operation in Eq. 2, action selection and action-value estimation are performed using two separate neural networks, as proposed by \citep{van2015deep}. The gradient of neural network is clipped at each training step to a value of 0.5 to avoid harmful oscillations on neural network parameters. Finally, the hard copy operation in the original DQN paper is replaced by a moving average copy to smoothen the training process. It is noteworthy that a number of other DQN improvements in the literature are investigated but since they provided small to no improvement on the results they are not reported here.

\subsubsection{Proximal policy optimization (PPO)} 
\citep{schulman2017proximal}, is a widely successful actor-critic method that builds on top of the policy gradient formulation to update its stochastic policy ($\pi_\theta$):

\begin{equation} \label{eq:3}
L_\theta^\pi=\mathbb{E}\left[(\pi_\theta (a_t|s_t ))/(\pi_{\theta_{old}} (a_t|s_t ) ) \hat{A}_t \right]
\end{equation}

where $\hat{A}_t$ is the advantage function and represents the difference between the value function of the selected action, $Q(s_t,a_t )$, and the average value function for that state over actions. Intuitively, maximizing Eq. \ref{eq:3} encourages the policy to increase the probability of action if the selected action performed better than average (i.e., the advantage is positive) and decrease the probability of relatively worse actions. However, this vanilla formulation tends to collapse the training process as taking large steps can easily move the policy into unrecoverable bad parameter spaces. To solve this issue, PPO restricts the ratio between current policy and previous policy by pessimistically clipping its value according to Eq. \ref{eq:4}:

\begin{equation}
\begin{aligned}
L_\theta^{\pi,clip} &= \mathbb{E}\left[ min(r_t(\theta) \hat{A}_t, clip(r_t(\theta), 1-epsilon, 1+\epsilon)\hat{A}_t) \right] \\
r_t &=  (\pi_\theta (a_t|s_t ))/(\pi_{\theta_{old} } (a_t|s_t ) ) 
\end{aligned}\label{eq:4}
\end{equation}

where $\epsilon$ determines the clipping range. Furthermore, PPO loss (Eq. \ref{eq:5}) has two additional terms in to train the advantage value $\hat{A}_t$ and to maximize the entropy ($L_\theta^Ent$) of the policy to encourage exploration.

\begin{equation} \label{eq:5}
L_\theta^{PPO}= \mathbb{E}\left[L_\theta^{\pi,clip}+c_1 L_\theta^A+c_1 L_\theta^{Ent} \right]
\end{equation}

PPO can only be trained using samples generated from its current policy (i.e., on-policy algorithm). This characteristic of PPO causes this algorithm to require a larger number of samples compared to off-policy algorithms where historic data can be reused for training the agent through the use of replay buffer. Empirically, the effectiveness of the PPO algorithm relies on collecting independent samples from multiple streams of environments often performed in parallel virtual environments.

\subsubsection{Soft actor critic (SAC)} 
\citep{haarnoja2018soft} is an off-policy actor-critic method that aims to maximize an alternate action-value function, called soft action-value, that considers not only the accumulative reward but also the entropy of its stochastic policy. Theoretically, soft action-value formulation encourages the agent to explore states with uncertain results. The soft action-value loss is presented in Eq. \ref{eq:6}:

\begin{dmath}\label{eq:6}
L_\theta^Q = \mathbb{E}_{s_t,a_t,s_{t+1},r_t,d_t \sim D}\left[ \left(min_{i=1,2}  Q_{\theta_i}(s_t,a_t) - r_t+\gamma(1-d_t)(min_{i=1,2} Q_{\overline{\theta}_i}^t (s_{t+1},a_{t+1} )-\alpha log(\pi_\phi (a_{t+1}|s_{t+1} )) ) \right)^2 \right]
\end{dmath}

where $\theta$, $\overline{\theta}$ and $\phi$ indicate the neural network parameters for the online action-value function, the target action-value function, and the policy respectively. The temperature parameter $\alpha$ determines the importance of entropy. SAC compensates for the overestimation of action-value functions by training two independent neural networks and taking the minimum of the two for loss calculations.

SAC policy loss is defined as the KL-divergence between the current policy and action-value softmax. To calculate the gradients of parameters through the stochastic node of policy sampling, the reparameterization trick is used. While the SAC only applies to environments with continuous action spaces, we developed a modified version of this algorithm that uses Gumble-softmax \citep{jang2016categorical} to perform the reparameterization for categorical action spaces (Eq. \ref{eq:7}).

\begin{equation} \label{eq:7}
L_\phi^\pi= -  \mathbb{E}_{s_t \sim D,\xi \sim G} \left[min_{i=1,2} Q_{\theta_i} (s_t,\tilde{a}_\phi (s_t,\xi))-\alpha log(\pi_\phi (\tilde{a}_\phi(s_t,\xi)|s_t )) \right]
\end{equation}

where $\xi$ is an independent noise sampled from a Gumble-softmax distribution and a $\tilde{a}_\phi$ is the reparametrized action. While the temperature parameter $\alpha$ can be potentially kept as constant, the SAC authors devise a formulation to simultaneously train this parameter in order to constrain it to a minimum target entropy $H$ (Eq. \ref{eq:8}).

\begin{equation} \label{eq:8}
L^\alpha= \mathbb{E}_{s_t \sim D,a_t \sim \pi} \left[-\alpha log(\pi_\phi (a_t|s_t)-\alpha H) \right]
\end{equation}

Although the above-mentioned three algorithms have inherent differences, we have tried to keep hyperparameters of the algorithms as consistent as possible. The hyperparameters of all algorithms are tuned to maximize the achieved score.

\section{Results and discussion}

\begin{figure}
	\centering
	\includegraphics[width=10cm]{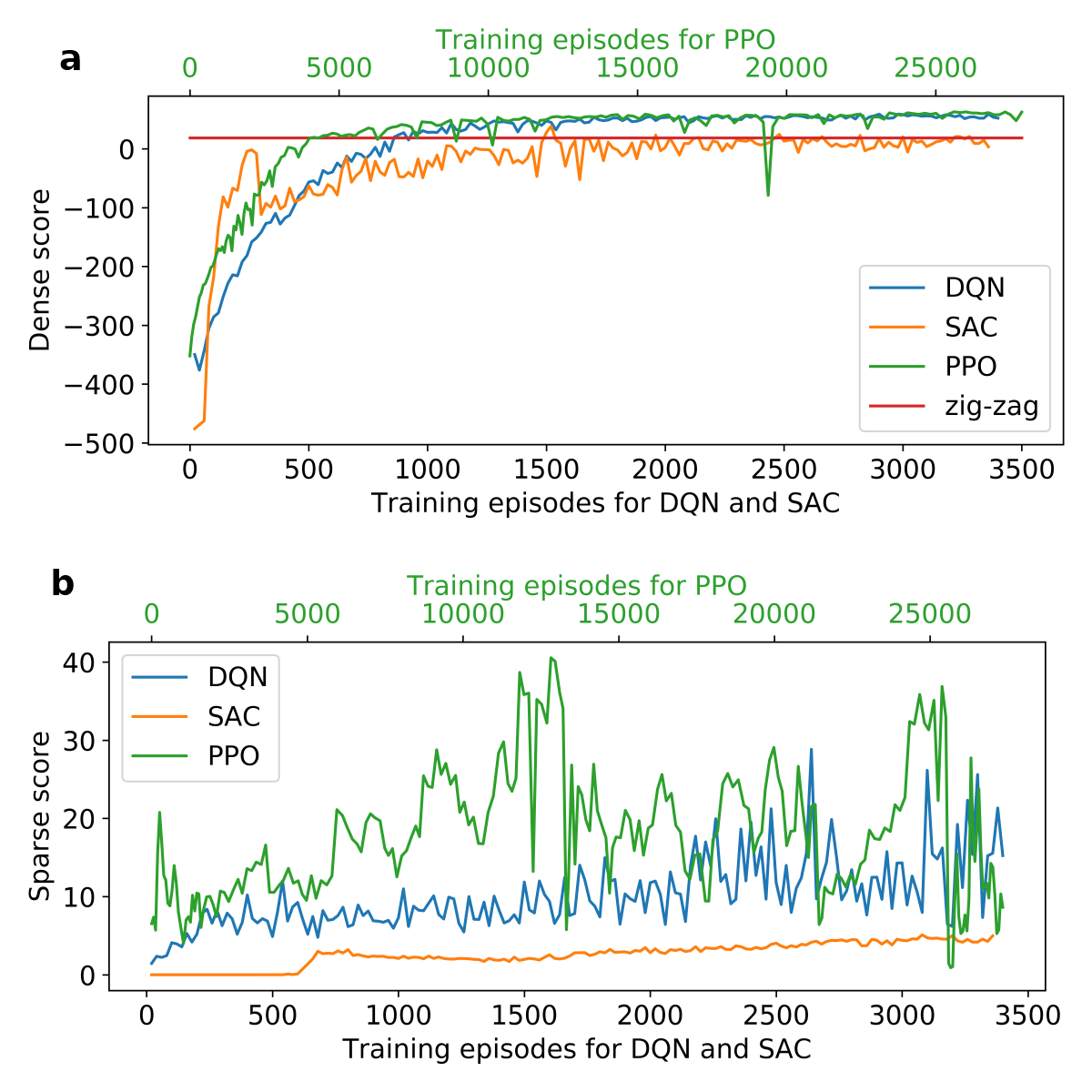}
	\caption{Learning curve of toolpath design system with the three DQN, SAC, and PPO algorithms for (a) dense and (b) sparse reward system. The horizontal axes for PPO results are plotted in a different scale (shown on top of each plot) from DQN and SAC results (shown on the bottom of each plot). As manual zig-zag toolpath strategy is plotted as a baseline for the dense reward system, however, such engineered solution does not apply for the sparse reward system.}
	\label{fig:fig3}
\end{figure}

We implemented the three discussed model-free algorithms with the proposed modifications using Tensorflow deep learning library. Each algorithm is trained to design the toolpath in two cases of dense and sparse reward structures as detailed in Sec. \ref{sec:alanysis_cases}. The learning curve of each algorithm is demonstrated in Fig. \ref{fig:fig3}a and b for dense and sparse reward structure, respectively. The reported score averages resulted scores for all training geometries from random initial nozzle location. Since the on-policy nature of the PPO algorithm requires far more episodes of toolpath generation than the two off-policy algorithms, the PPO results are plotted on a different horizontal scale for the number of episodes.

\begin{figure}
	\centering
	\includegraphics[width=15cm]{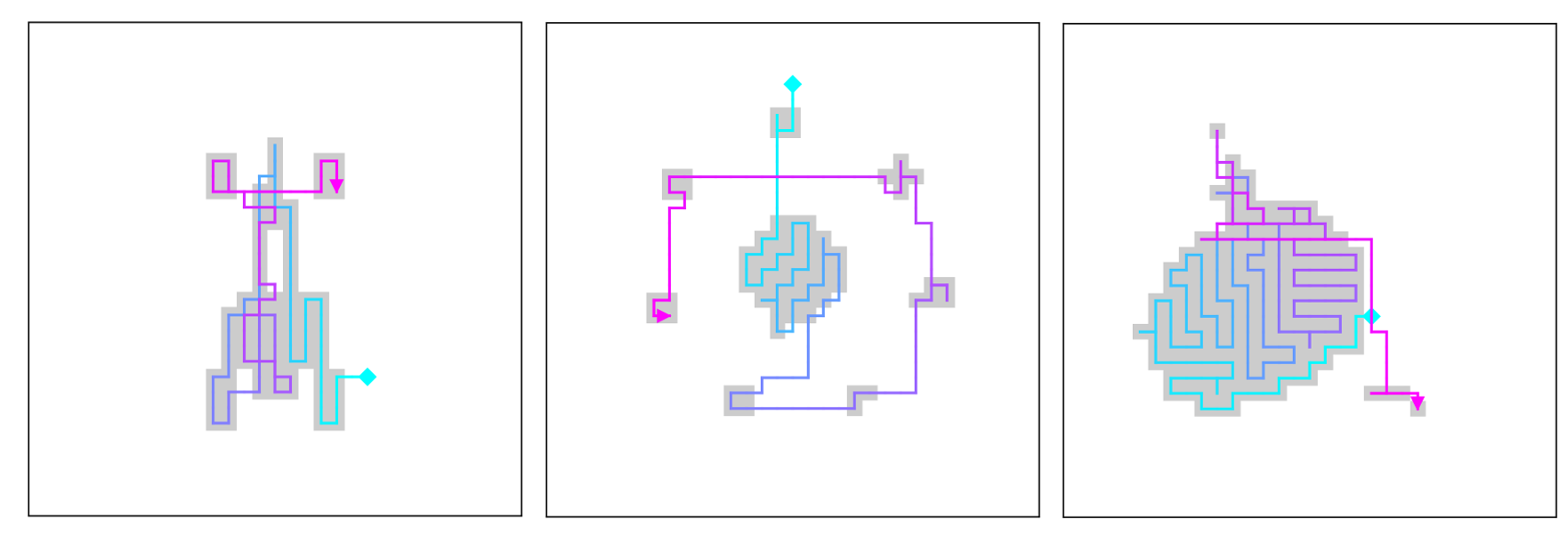}
	\caption{Three samples of the designed toolpaths by the trained PPO algorithm for random sections and starting locations. The section is depicted in light grey. The toolpath motion starts from the blue diamond shape, following a color gradient ending in a pink arrow shape.}
	\label{fig:fig4}
\end{figure}

As can be seen from Fig. \ref{fig:fig3}a while three algorithms gradually learn to improve their toolpath designs, SAC achieves a notably inferior performance. DQN and PPO reach a close performance easily surpassing a manually coded zig-zag toolpath. While the final performance of the PPO algorithm is 3 scores higher than DQN, DQN reaches a stable solution using 10 times less samples.
For the sparse case (see Fig. \ref{fig:fig3}b), the score achieved by PPO algorithm surpasses the two other algorithms, and similar to the previous case, SAC results in the worst performance. The highest score of the algorithms for the two cases is reported in Table \ref{tab:table} and three samples of the designed toolpaths with trained PPO algorithm is demonstrated in Fig. \ref{fig:fig4}.

\begin{table}
	\caption{Highest score of model-free algorithms for two reward structure cases.}
	\centering
	\begin{tabular}{lll}
		\toprule
		%\multicolumn{2}{c}{Part}                   \\
		%\cmidrule(r){1-2}
		Algorithm     & Dense reward     & Sparse reward \\
		\midrule
		DQN     & $59.31$	          & $28.83$    \\
		PPO     & $62.98$	          & $40.54$     \\
		SAC     & $38.71$	          & $5.11$  \\
		\bottomrule
	\end{tabular}
	\label{tab:table}
\end{table}

Our results show that model-free reinforcement learning is a feasible approach for high-dimensional manufacturing design systems, such as toolpath design tools, especially if a dense reward system exists or it is feasible to engineer such a feature by breaking the task into meaningful step-by-step reward increments. DQN-based algorithms show great potential in this realm as they offer decent accuracy and sample efficiency. Although SAC algorithm is reported to produce state-of-the-art benchmarks in many robotics tasks, it is incapable of handling the intricacies of toolpath design.

In the case of a sparse reward structure, investigated model-free approaches struggle to optimize the toolpath. PPO progressively learns better solutions; however, its excessive on-policy sample requirement makes this algorithm only applicable to cases where a robust simulation of the physics exists. Novel combinations of model-free and model-based solutions are necessary to solve this class of problem on experimental data.

\section{Conclusions}
In conclusion, we proposed a new framework for the toolpath design of metal-based additive manufacturing processes by formulating a reinforcement learning problem. Modified versions of three state-of-the-art model-free RL algorithms are used to develop toolpaths on a virtual additive manufacturing environment, and our results indicate that model-free RL algorithms achieve high scores of the tasks especially in existing of dense reward structures.

\section{Acknowledgements}
This work was supported by U.S. Department of Commerce
445 (70NANB19H005) and National Institute of Standards and Technology as part of the Center for Hierarchical Materials Design (CHi-MaD).

\bibliographystyle{unsrtnat}
%\bibliography{references}  %%% Uncomment this line and comment out the ``thebibliography'' section below to use the external .bib file (using bibtex) .





% \end{thebibliography}

\end{document}